# Unsupervised Features for Facial Expression Intensity Estimation over Time

Maren Awiszus, Stella Graßhof, Felix Kuhnke, Jörn Ostermann
Institut für Informationsverarbeitung
Leibniz Universität Hannover
www.tnt.uni-hannover.de

## Abstract

*The diversity of facial shapes and motions among persons is one of the greatest challenges for automatic analysis of facial expressions. In this paper, we propose a feature describing expression intensity over time, while being invariant to person and the type of performed expression. Our feature is a weighted combination of the dynamics of multiple points adapted to the overall expression trajectory. We evaluate our method on several tasks all related to temporal analysis of facial expression. The proposed feature is compared to a state-of-the-art method for expression intensity estimation, which it outperforms. We use our proposed feature to temporally align multiple sequences of recorded 3D facial expressions. Furthermore, we show how our feature can be used to reveal person-specific differences in performances of facial expressions. Additionally, we apply our feature to identify the local changes in face video sequences based on action unit labels. For all the experiments our feature proves to be robust against noise and outliers, making it applicable to a variety of applications for analysis of facial movements.*

## 1. Introduction

Automatic analysis of facial expressions provides a number of challenges including recognition of basic emotions, detection of facial action units (AU) [3] and the creation of statistical face models. However, if multiple persons are asked to perform the same expression, the recordings differ not only in length but also in appearance. Even two recordings of the same person performing the same expression will likely result in noticeable variations. It is still unclear how different recordings of a moving face can be compared. To address this problem, we propose a method to estimate the facial expression intensity over time, which is independent of the type of expression, its magnitude, and the person performing it.

In this work, we show that our method for expression intensity estimation performs better than a current state-of-the-art approach [20] on the BU4DFE [19] dataset. However, our additional experiments highlight the general applicability to other face analysis problems. We are able to temporally align sequences of 3D dynamic facial expressions with our estimated expression intensities. Temporal and spatial alignment of data is a crucial step for statistical model generation, especially for human faces and bodies [4, 5, 6].

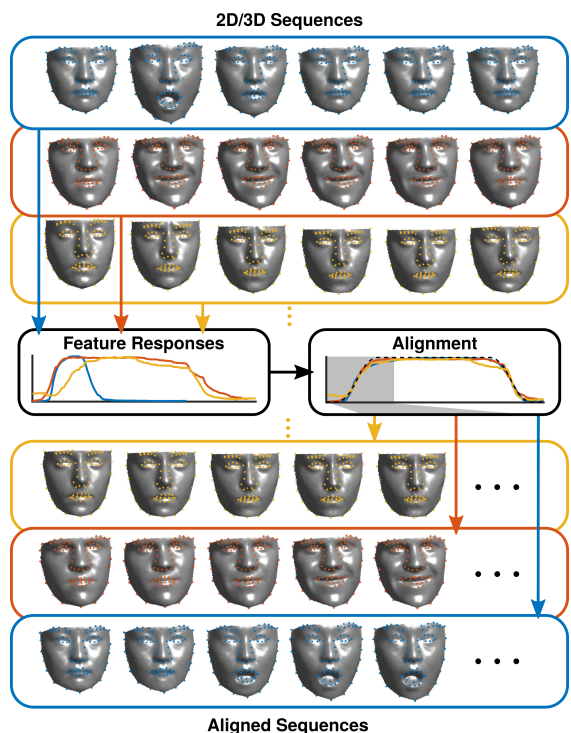

Figure 1. Pipeline for one of the applications of our response: Temporal alignment of multiple sequences of 2D or 3D feature points, whereas each is represented by one of our proposed one-dimensional feature responses. (Details in Section 3.3. Figure is best viewed in color.)

With our proposed weighting of feature points, we show that since different persons perform the same emotion very differently, a clustering of these emotion labels is near impossible by revealing the subclusters within the labeled emotion classes. Furthermore, our method is applicable to facial action unit (AU) intensity and the proposed weighting reveals activation patterns of specific landmarks corresponding to those AUs.

**Related Work** Automatic recognition of facial expressions of emotions is an active research topic. The interested reader might refer to [13] or [1] for recent surveys on this topic. However, in this work our goal is to provide a feature for relative facial expression intensity over time not specific to a certain emotion or action.

The most similar and recent work compared to our approach is described in [20], where the authors use landmarks, local binary patterns (LBP), and wavelet coefficients, combined with a regression model to describe and estimate expression intensities over time. The major differences to our work are that our approach is model-free and unsupervised, in a sense that we do not need any annotations and that our proposed feature can be directly retrieved from facial landmarks [15]. Also we found that we clearly outperform [20] on the BU4DFE database.

Another application for our method is using the expression intensity responses for a temporal alignment. This can be useful for different applications, such as statistical models, e.g. [4]. Since the quality of the alignment is crucial for the quality of resulting statistical models, we evaluate the proposed features by temporally aligning sequences of 3D facial feature points of the BU4DFE database [19].

In [10] the authors describe a similar feature to temporally synchronize videos of facial performance. While they also use local PCA on 2D landmarks to compute a cost matrix between frames of two sequences, they do not assign individual weights to the individual facial feature points. Though the cost matrix takes use of audio and visual information for alignment, it must be processed to deal with ambiguities.

The commonly used algorithms for alignment of pairs of sequences, such as Dynamic Time Warping (DTW) [12], depend on a one-dimensional representation of the signals to be aligned. In [21], the authors introduce an extension of DTW, the Generalized Canonical Time Warping (GCTW), enabling the use of multi-dimensional representations for the alignment. A common method for generating representative features is performing a principal component analysis (PCA) over all points. The first principal components then follow the course of the highest variances in the points and are used for further alignment and analysis. For example in [21], the first three components are used to align multiple sequences of moving mouths by GCTW. Thereby, it is assumed that the first principal component corresponds to the motion to be aligned. However, especially for falsely tracked or noisy points, this is not the case. To remove such erroneous points extensive preprocessing would be required before calculating PCA, while our proposed method discards them automatically. Instead of using PCA, in [14] the authors suggest features based on 2D landmarks and their differences to detect AUs and their temporal phases. To select the most influential features among 840 features per frame, they train binary classifiers. In contrast to that, our method requires no training and can also be used on 3D Points.

In [2] the authors describe time-varying 3D facial feature points by topological and geometric descriptors. However, in contrast to our approach, the utilized features are predefined for a small set of facial points, and e.g. do not allow to distinguish an open mouth or eye from a closed one.

To summarize, the contributions of this work are as follows:

- A novel method to estimate relative facial expression intensity over time.
- Local features are automatically weighted, which provides a meaningful ranking and robustness against outliers.
- We outperform state-of-the-art method [20] for expression intensity estimation on the BU4DFE [19].
- We reveal subclusters of facial feature points for each emotion, indicating person-specific performance.
- AU-specific facial feature points can be identified automatically.
- We provide temporal alignment information for the BU4DFE database [19][1].

The remainder of the paper is organized as follows: In Section 2, we explain our proposed feature and how it can be used for temporal alignment. Section 3 contains the performed experiments, namely comparison to the method of [20], temporal alignment, detection of emotion subclusters, and action unit analysis. In the final section, we summarize and discuss our paper.

## 2. Proposed algorithm

In the following we assume the input data sets share a similar general temporal behaviour, which is described in section 2.2.

We assume our input data to consist of $k = 1, \cdots, S$ sequences with $i = 1, \cdots, N$ $d$-dimensional feature points. We define $T_k$ as the number of frames of sequence $k$ and $t = 1, \cdots, T_k$ as the corresponding index to address each frame. Each sequence $k$ can be described as a 3-mode tensor $\mathcal{F}_k \in \mathbb{R}^{d \times N \times T_k}$. Our goal is to reduce the dimensionality of each three-dimensional tensor $\mathcal{F}_k$ to a one-dimensional representation with $T_k$ values, which is our final response representing the expression intensity over time. These $S$ sequences can then be aligned to all have length $T$. This allows to sort the data into one tensor $\mathcal{F} \in \mathbb{R}^{d \times N \times T \times S}$ which can then be used for a statistical model.
In the following we describe the dimensionality reduction step by step, leading to our proposed feature for expression intensity. Furthermore, we describe how it can be used to align sequences which we will apply in experiment 3.3.

---

[1] www.tnt.uni-hannover.de/project/facialanimation/bu4dfe-alignment

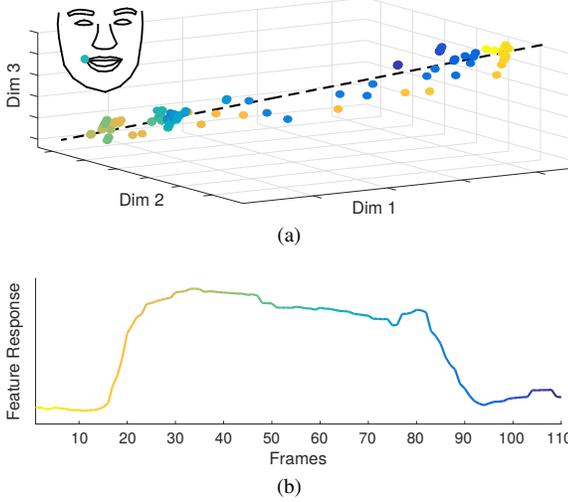

Figure 2. (a) The coordinates of a tracked point, namely the corner of the mouth during a sequence depicting happiness, over time with their colors corresponding to the frames in the sequence starting at yellow and ending at blue. The dotted line depicts the axis of the first eigenvector. The curve in (b) illustrates $\mathbf{R}_k(i,:)$, which corresponds to the distances of the points projected onto the first eigenvector to the location of the point in the first frame.

In this paper, we denote row $i$ of a matrix $M$ as $M(i,:)$ and column $j$ as $M(:,j)$.

## 2.1. Calculating response matrix $\mathbf{R}_k$

First, global movement for each frame is eliminated and each face shape is translated, such that the base of the nose lies in the origin. Then, we reduce each $d$-dimensional point $i$ to a 1-dimensional representation, by calculating PCA for the matrix $\mathcal{F}_k(:,i,:) \in \mathbb{R}^{d \times T_k}$. We then project the $d$-dimensional columns of the matrix onto the first eigenvector, and calculate the distance between each column and the first along that axis, which results in the local responses $\mathbf{R}_k(i,:) \in \mathbb{R}^{T_k}$. Fig. 2 shows one example response. Calculating this for each point results in the full matrix $\mathbf{R}_k \in \mathbb{R}^{N \times T_k}$ in which we now have $N$ possible representations for our sequence. In the next sections we describe how the quality of each feature response $\mathbf{R}_k(i,:)$ can be defined, leading to a ranking for each corresponding feature point $i$.

## 2.2. Calculating approximated response $\mathbf{r}_k^A$

To define the quality for each of the $N$ responses per sequence, we assume one specific course of the sequence to find the points that follow it most closely. This response may already be given or calculated by other means, as in [20], in which case the steps in this subsection can be skipped. An example for the case with provided annotations is given in section 3.5.

In the following, we assume the faces in the sequences start in a neutral expression, then change to a different one (e.g. happiness) in full extend with highest intensity and then return to neutral again. We assume all parts of the face to change synchronously. This means our approximated response $\mathbf{r}_k^A \in \mathbb{R}^{T_k}$ can be described as a box function:

$$\mathbf{r}_k^A = \begin{cases} 0, & \text{for} \quad 0 \leq t \leq t_{k,1} \\ 1, & \text{for} \quad t_{k,1} < t < t_{k,2} \\ 0, & \text{for} \quad t_{k,2} \leq t \leq T_k \end{cases}, \quad (1)$$

where 0 represents the neutral and 1 the full expression. Assuming each sequence can be represented by one $\mathbf{r}_k^A$, they only differ in the frames $t_{k,1}$ and $t_{k,2}$ where the changes of facial expression occur. However these are unknown. To find these frames for each sequence, we use a feature response representing the derivation over time, where the extremes correspond to $t_{k,1}$ and $t_{k,2}$. We calculate the derivation $\mathbf{R}_k^\delta$ of all responses $\mathbf{R}_k(i,:)$ with a convolution.

This results in $N$ derivative responses, whose absolute values are merged by calculating the median of each column $t$ as follows:

$$\mathbf{r}_k^\delta(t) = \text{median}\left\{|\mathbf{R}_k^\delta(:,t)|\right\}. \quad (2)$$

We use the median since it is robust against outliers, which can be a problem as will be described in section 3.3. In the end, we obtain one derivative response $\mathbf{r}_k^\delta \in \mathbb{R}^{T_k}$ per sequence. In Fig. 4(a)-(d) this derivative response is illustrated in green, showing two distinct maxima. We calculate these maxima and set $t_{k,1}$ and $t_{k,2}$ to the frames of the first and second maximum. Using $t_{k,1}$ and $t_{k,2}$ with Eq. (1) we obtain one approximated response $\mathbf{r}_k^A$ of for each sequence, depicted in yellow in Fig. 4(a)-(d).

## 2.3. Ranking and final response $\mathbf{r}_k$

Given the approximated response $\mathbf{r}_k^A$ per sequence, we define a ranking of our $N$ local responses $\mathbf{R}_k(i,:)$ by calculating the distance between them and the approximated response $\mathbf{r}_k^A$.

As the eigenvector and therefore the local response can be mirrored in the second dimension, $\mathbf{R}_k(i,:)$ can follow the approximated course in the negative direction. To account for this, we need to calculate the distance in both directions. Furthermore, the responses $\mathbf{R}_k(i,:)$ are scaled to [0,1] to make them comparable to $\mathbf{r}_k^A$. We therefore define $\widehat{\mathbf{r}}_{k,i}$ as $\mathbf{R}_k(i,:)$ scaled to [0,1] and $\widecheck{\mathbf{r}}_{k,i}$ as $-\mathbf{R}_k(i,:)$ scaled to [0,1]. With this we define our subsequently used response:

$$\widetilde{\mathbf{R}}_k(i,:) = \begin{cases} \widehat{\mathbf{r}}_{k,i}, & \text{if } \left\|\mathbf{r}_k^A - \widehat{\mathbf{r}}_{k,i}\right\|_2 < \left\|\mathbf{r}_k^A - \widecheck{\mathbf{r}}_{k,i}\right\|_2 \\ \widecheck{\mathbf{r}}_{k,i}, & \text{otherwise} \end{cases} \quad (3)$$

Given the adapted response $\widetilde{\mathbf{R}}_k(i,:) \in \mathbb{R}^{T_k}$ the distance to the approximated response is:

$$D_{k,i} = \left\| \mathbf{r}_k^A - \widetilde{\mathbf{R}}_k(i,:) \right\|_2 \quad (4)$$

Considering the point corresponding to the lowest distance resembles the approximated response $\mathbf{r}_k^A$ best and thereby represents the course of the sequence $k$ best, its influence should be high. To encode the contribution of each point $i$ to the final response (expression intensity), the point-wise distances are converted to weights, scaled between $[0, 1]$ as:

$$W_{k,i} = 1 - \frac{D_{k,i}}{\max\{D_{k,i} | i = 1, \ldots, N\}}, \quad (5)$$

leading to the final response:

$$\mathbf{r}_k(t) = \sum_{i=1}^{N} W_{k,i} \cdot \widetilde{\mathbf{R}}_k(i, t). \quad (6)$$

Some examples of the one-dimensional final response, representing the relative expression intensity, are shown in Fig. 1 in different colors for the examples, and in Fig. 4 in orange. Please note that $\mathbf{r}_k$ does not depend on the person or expression. In fact, the response depends on the approximated response, which can be created automatically from data as in experiment 3.3 and 3.4 or from human annotations as in experiment 3.5.

### 2.4. Alignment

Given a one-dimensional representation of each sequence, GCTW [21] can be used to align multiple sequences at once. Unfortunately the simultaneous alignment does not allow to incorporate prior knowledge of the properties of the sequences to be aligned, e.g. the exact frames of the transitions from neutral to expression or the desired length of the sequence.

However, the used databases provide prior information, enabling the assumption that a known template response[2] $\mathbf{r}^T$ is provided. We therefore choose to perform pairwise alignments and additionally found that computing these only takes a fraction of the time compared to a simultaneous alignment. In our experiment 3.3 we used a smoothed trapezoid as $\mathbf{r}^T$, which is illustrated as dotted line in Fig. 1 (see 3.3 for more details).

## 3. Experiments

In the following section, we evaluate our method on four tasks: expression intensity estimation, temporal alignment of expression sequences, identifying the most influential facial feature points for an action unit, and discovery of subclusters within groups labeled with the same emotion. We show results of our algorithm on using two different databases. We discuss difficulties encountered with the databases and show that our approach outperforms other methods for the same tasks.

---

[2]Please note that the approximated response $\mathbf{r}_k^A$, the final response $\mathbf{r}_k$ and the template response $\mathbf{r}^T$ differ.

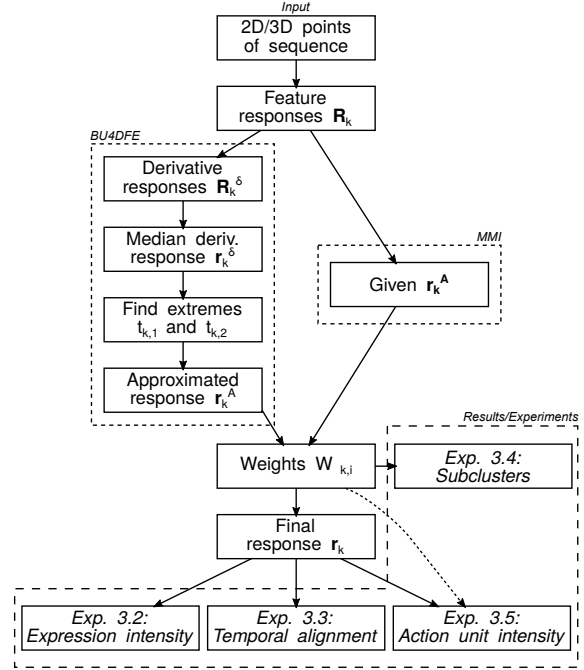

Figure 3. Summary of the algorithm described in Section 2 With indications which part of it is used in which experiment.

To ease the navigation through our diverse experiments we provide a schematic overview of our method and the experiments in Figure 3.

### 3.1. Datasets

#### 3.1.1 BU4DFE

The BU4DFE database [19] contains $S = 606$ sequences with different lengths from 101 persons, performing the same 6 prototypical emotions: anger, disgust, fear, happiness, sadness, surprise. For each frame $N = 83$ facial features points in $d = 3$D are provided, such that each sequence $k$ can be ordered into a data tensor $\mathcal{F}_k \in \mathbb{R}^{d \times N \times T_k}$, $k = 1, \ldots, S$. While each sequence is supposed to start with a neutral expression, then change to one emotion with full expression intensity, and return to neutral, the actual course may vary. We discuss this in Section 3.3.

#### 3.1.2 MMI

The MMI database [11] consists of photos and videos of 19 subjects performing different expressions and specific AUs. For our experiments, we only use frontal view videos of those persons which had at least one annotated AU resulting in a subset of 302 sequences of 16 persons. The labels for neutral (NE), onset (ON), apex (AP) and offset (OF) of the AUs in those sequences are supplied. Like the BU4DFE, the faces in the sequences start neutral, change to an expression or a facial AU, and then change back to neutral. Unlike the BU4DFE, this dataset does not contain tracked feature points. We used the dlib-toolbox [9] to generate 68 2D landmarks for each frame. (so $N = 68$ and $d = 2$)

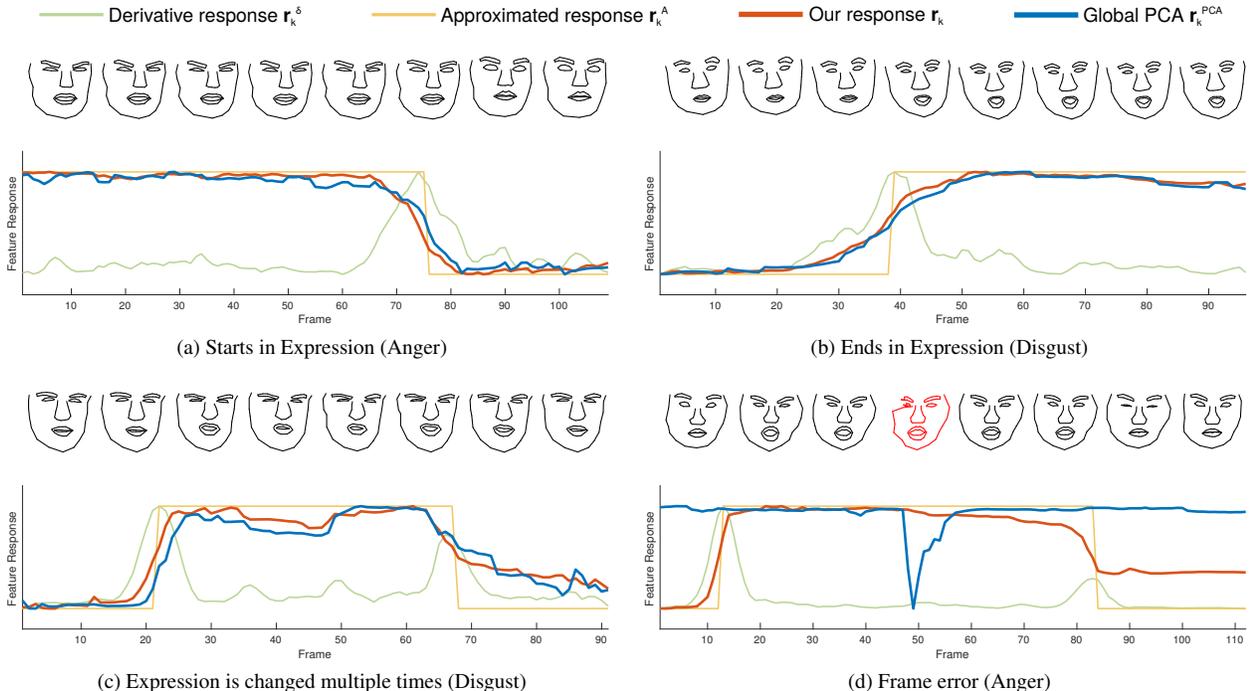

Figure 4. Examples for the problems encountered in the database: (a) A sequence already starting in the expression, (b) a sequence ending in the expression, (c) a person changing their expression multiple times (note the mouth around frame 40), (d) a person with a distorted frame (frame 50).

### 3.2. Expression intensity estimation on BU4DFE

In contrast to emotion classification, which tries to find the emotion label of a given image or frame of a sequence, emotion intensity estimation is the task of estimating the intensity or strength of a given emotion. Given the sequences of the BU4DFE database [19], we apply the proposed algorithm described in Sec. 2 to estimate the expression intensity over time $\mathbf{r}_k$ (Eq. (6)) for each sequence $\mathcal{F}_k \in \mathbb{R}^{d \times N \times T_k}$ and compare our results to [20].

In [20], expression intensities over time and therefore responses similar to ours are calculated and evaluated on the BU4DFE dataset. In the following we describe important differences between our method and theirs as well as show our performance on the same and extended experiments.

Zhao *et al.* [20] create a pseudo ground truth for BU4DFE based on manual labels by assuming that the ground truth expression intensity can be represented by a triangular function. It essentially describes a linear change to and from the apex. In contrast to that in Sec. 3.3 we assume a trapezoid, which we argue to better represent the motions of BU4DFE. However, for the sake of comparison in this experiment we use the triangular function proposed by [20]. Also, they used a reduced set of 120 sequences instead of the full 606, which is probably due to problems with some sequences of the database, which are further described in Sec. 3.3. In contrast to that, our method can be applied to all 606 sequences as we will show in the results. Furthermore, as already stated in Sec. 1, our method is completely unsupervised as we only rely on the facial landmarks with no additional features, like LBP and Gabor wavelet coefficients, which are used in [20], and we do not need to train a regression model at all.

In order to compare their method, called ordinal support vector regression (OSVR), to other methods, Zhao *et al.* used three metrics: mean absolute error (MAE), Pearson correlation coefficient (PCC), and intra-class correlation coefficient (ICC). They compared their method to two unsupervised methods, ordinal regression (OR) [7] and Rankboost [18]. We found that part of the code and results for [20] are publicly available and therefore the method of calculating the three metrics will be the same as in [20]. The only information we are missing to be unassailably comparable is which 120 of the 606 sequences were used to calculate the results. In order to still be reasonably comparable, we manually selected 120 sequences which exhibited none of the problems described in 3.3. In these sequences, we manually labeled one frame as the apex in order to faithfully recreate the conditions in [20]. Additionally, we calculated the metrics for the entire set of 606 sequences with automatically generated apex frames for each response. The frames chosen as the apex were the halfway point between the calculated $t_{k,1}$ and $t_{k,2}$ from Eq. 1. The results for both experiments are shown in table 1.

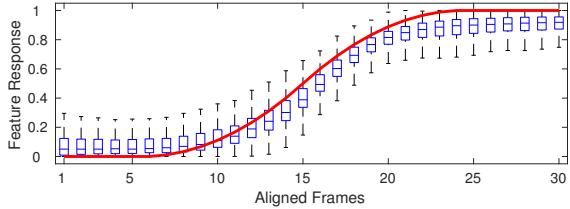

Figure 5. This figure shows the alignment of one transition for each of the 606 sequences via a boxplot. The red line represents our template response $\mathbf{r}^T$ and the boxes represent the distribution of expression intensities for their representative frame. Outliers outside of the 95% confidence interval were omitted.

|  | MAE | PCC | ICC |
| --- | --- | --- | --- |
| OR [7] | 4.7803 | 0.5306 | 0.0850 |
| Rankboost [18] | 5.3637 | 0.4067 | 0.2390 |
| OSVR-L1 [20] | 2.5736 | 0.5393 | 0.4808 |
| OSVR-L2 [20] | 2.2424 | 0.5453 | 0.5025 |
| Ours | **1.354** | **0.9354** | **0.9013** |
| Ours (606 seq.) | 1.476 | 0.9125 | 0.8765 |

Table 1. Comparison of expression intensity estimation results on BU4DFE [19]. The metrics used are the mean absolute error (MAE), Pearson correlation coefficient (PCC), and intra-class correlation coefficient (ICC). These metrics are calculated as described in [20]. Our method outperforms state-of-the-art methods even if all sequences of the BU4DFE are used for the calculation.

Please note that the values for OR [7] and Rankboost [18] are taken from [20].

Our method clearly outperforms the other methods, even if we use all 606 sequences.

### 3.3. Temporal alignment of 3D face scans from BU4DFE

The goal of the following experiment is to compute a temporal alignment of a set of sequences with varying length, thereby unifying the length and enabling to sort the data into one data tensor or matrix for various applications, e.g. statistical analysis or face model estimation [4]. However, for a statistical face model we are only interested in one transition from neutral to apex.

As a result of the previous experiment, one-dimensional expression intensities over time have been calculated for each of the 606 3D sequences of the BU4DFE database. We now use the proposed algorithm to temporally align the sequences as described in Sec. 2.4. An overview of the process is illustrated in Fig. 1 for three sequences, where for each sequence facial feature points are plotted onto the dense meshes, along with their unaligned and aligned feature responses.

As described previously, we assume that each sequence starts with a neutral expression, changes to full expression (with highest intensity) and returns to neutral. To align the sequences we define a template response $\mathbf{r}^T$, such that the first and last 30 frames of the aligned sequences contain the transitions from and to the neutral expression. The template response is derived from the median of the previously calculated responses $\mathbf{r}_k$ over all sequences leading to a smoothed trapezoid that was adjusted for symmetry. The first 30 frames and therefore the first transition of $\mathbf{r}^T$ can be seen as the red curve in Fig. 5.

As the transition between neutral and full expression contains the frames we are interested in for this experiment, we should be able to just sample the first 30 frames of the sequences aligned to $\mathbf{r}^T$ for that. However, some sequences exhibit problems that need to be addressed first:

*Problem 1:* In some of the sequences, the person either already starts in a facial expression or does not return to neutral, technically leaving one half of the assumed facial motion undefined. This can be seen in Fig. 4(a) and (b). However, adjusting the approximated response $\mathbf{r}_k^A$ of Eq. (1) to only include one half, i.e. either $t_{k,1}$ or $t_{k,2}$, the final feature response $\mathbf{r}_k$ can still be calculated. Therefore, at least one of the two transitions can be aligned to the template response $\mathbf{r}^T$.

After the alignment, we checked both transitions for each sequence, i.e. the first and last 30 frames, and calculated the distance from the aligned response $\mathbf{r}_k^{alig}$ to the template response $\mathbf{r}^T$. We then select the transition with the lower alignment error. Should that transition be the second one (from expression to neutral), we flip the order of the shapes so that the transition is again from neutral to the expression. This guarantees that at least one of the two transitions is aligned for all sequences.

*Problem 2:* There are sequences with more than two transitions. An example of such a sequence is shown in Fig. 4(c), where the mouth shape changes more than two times. One advantage of our method is that though multiple transitions may be present, we are still able to retrieve one alignment, based on one of these transitions, though we did not consider which of the transitions is more relevant.

*Problem 3:* The last problem we observed in the database are sequences with severe tracking errors. These may be short deviations for a few frames, as one example shows in Fig. 4(d), or a recurring jumping of one point between different locations. In these cases, the proposed re-sponse can still represent the underlying expression, while the global PCA (described in more detail below) does not. This is due to the fact that the response resulting from PCA follows the highest variance, which is influenced by the tracking errors.

As there are some sequences where only one transition is present, either from neutral to expression or vice versa, we chose to calculate a temporal alignment for one direction only. Based on the assumptions that: (1) there is at least one transition in each sequence and (2) we are able to mirror the direction if needed, we can guarantee a meaningful temporal alignment from neutral to full expression. The described alignment procedure allows sampling of an arbitrary number of frames from the aligned transitions, thereby enabling the creation of a tensor (or matrix) and corresponding model.

The boxplots based on all aligned sequences in Fig. 5 illustrate that the sequences are well aligned to the template, depicted in red. In the following we provide an additional quantitative comparison.

**Global PCA:** To compare our method for temporal alignment to another commonly used one, we compute the proposed feature using the global PCA as in [21]. When calculating the global PCA response, the same problem as described in Section 2.3 applies, i.e. the response can be mirrored along the intensity axis. To make the results comparable, we adjusted the response similar to Eq. (3), substituting our final response from the proposed algorithm for $\mathbf{r}_k^A$.

As there is no ground truth alignment provided for the BU4DFE database, we quantify the quality of the alignment by the frame-wise error between the aligned and the template response over all sequences for the 30 previously described aligned frames with the mean squared error:

$$\text{MSE} = \frac{1}{S} \sum_{k=1}^{S} ||\mathbf{r}_k - \mathbf{r}^T||_2^2, \quad \mathbf{r}_k, \mathbf{r}^T \in \mathbb{R}^{30}. \quad (7)$$

In this case $\mathbf{r}_k$ stands for either the proposed response or the one calculated with global PCA [21].

The global PCA feature response results in a MSE of 1.35 while the MSE for our proposed response is only 0.48. The main reason for this are faulty sequences like Fig. 4(d) which can not be aligned correctly for the global PCA. This proves that our feature response is more robust against outliers and noise with a high variance compared to the global PCA.

### 3.4. Person-specific subclusters of emotions

The weights defined in Section 2 measure how close a point follows one facial expression, thereby offering a visualization of which points actually contribute to one facial expression. Using the weights $W_{k,i}$, we investigated the different ways people express the 6 prototypical emotions by clustering each of them into 3 clusters with a hierarchical tree using the Ward distance [8]. We found 3 clusters to represent our findings best. Figure 6 shows the mean shapes of the resulting clusters, where each sequence is represented by one shape with the full expression. Additionally, the 83 landmarks used are plotted in red with the size of each point corresponding to their weights $W_{k,i}$.

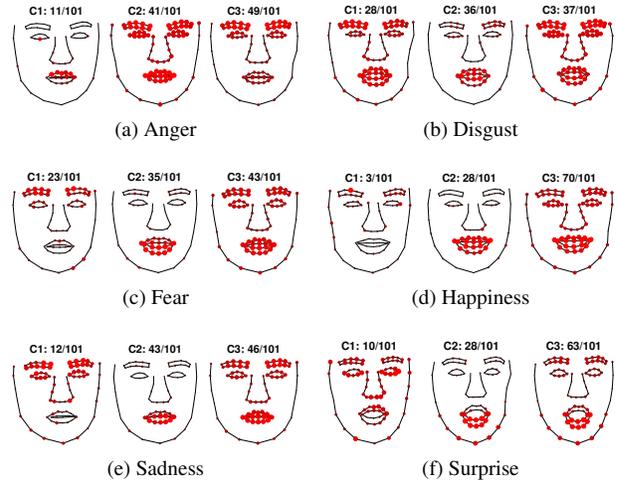

Figure 6. Subclusters for the six prototypical emotions, clustered with a hierarchical tree. The faces represent the mean shapes of their respective clusters, the numbers above them indicate how many belong to that cluster. The size of the red points corresponds to the mean of the weights $W_{k,i}$ for their respective cluster, therefore indicating which parts of the face are most important.

The results show, that different persons perform the same expression very differently. Mostly, people can be clustered as relying either on their mouth, eyes and eyebrows, or both regions to express emotions. This shows that, looking only at the landmarks, it is a very complex and maybe impossible task to cluster or classify the six emotions flawlessly.

### 3.5. Identifying most influential points of action units

In the following section, we describe how our proposed algorithm can be used to calculate responses for specific facial action units (AUs) [3] and how the resulting weights can be used to find feature points that are descriptive for AUs.

To **generate AU-specific responses**, the provided frame numbers for the labels NE, ON, AP and OF (see 3.1.2) are used to generate approximated responses for each AU. This is done by assigning the value zero to the response for the frames between NE and ON, setting it to 1 for frames between AP and OF, and interpolating linearly between 0 and 1 from ON to AP and OF to NE. These approximated responses $\mathbf{r}_k^A$ can be used in our algorithm to compute the final response $\mathbf{r}_k$. In the following, the lowest $\frac{3}{4}$ of the weights $W_{k,i}$ are set to 0, resulting in 17 remaining weights.

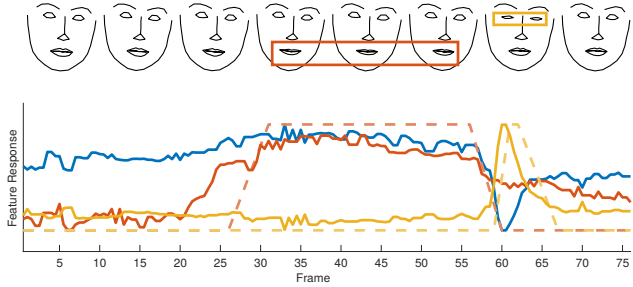

Figure 7. One example sequence from the MMI database [11], with given labels for onset, apex offset and neutral for two action units (AU17 and AU45). Using these, we define one approximated response (orange and yellow dashed lines) for each AU, which we then use to calculate the final responses (solid lines). In contrast to that, the response of global PCA (solid blue line) is not able to disentangle both AUs.

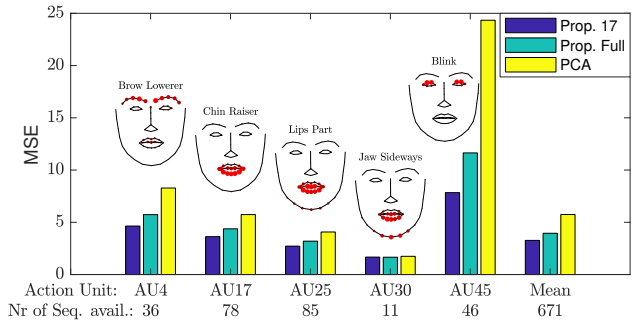

Figure 8. The bars illustrate the MSEs between the different feature responses and their corresponding approximated responses for specific AUs. The face shapes above the bars visualize the mean weights for the AUs.

In Fig. 7 different responses for one sequence with chin raise (AU17) and blinking (AU45) are shown, illustrating that the proposed 1-dimensional feature response is capable of finding even small movements in tracked points. Please note that the responses in Fig. 7 are less smooth than the ones shown in Fig. 4. This is caused by the single-frame landmark detection approach which introduces noise. However, the proposed framework seems to be robust against this noise for most AUs. We are able to identify AUs even with very short activation patterns as depicted in Fig. 7.

**Analysis of specific AUs:** In the selected subset of 302 sequences, there are 671 cases with a full label set (NE-ON-AP-OF-NE) for 36 different AU labels. For each of these, the mean squared error between the approximated response $\mathbf{r}_k^A$ and the corresponding final response $\mathbf{r}_k$ was calculated. This is done for the global PCA, our algorithm with all weights kept and with only the 17 highest weights. A bar plot for selected AU labels is provided in Fig. 8. Additionally a bar plot for all individual AU labels can be found in the supplementary material. The mean of those errors for the three types of responses are $MSE^{\text{PCA}} = 5.74$, $MSE^{\text{Prop. full}} = 3.95$ and $MSE^{\text{Prop. 17}} = 3.27$. The proposed feature in both cases has a lower error than the global PCA. The difference stems from the fact that our weighting lowers the influence of unrelated points which still contribute to global PCA. Setting the lower weights to zero minimizes that influence even further, resulting in an even lower error.

We observed that the difference between PCA and the proposed feature is particularly high for AU45 (Blinking), as can be seen in Fig. 7 and Fig. 8. This is a result of that AU rarely being the principal direction of the sequence which means the global PCA does not capture these small movements well. In cases in which the difference between the MSE values is low, we observed the corresponding AUs generally being of a high magnitude so the global PCA feature follows it very well, for example AU30 (Jaw left/right).

Additionally, similar to Section 3.4, the highest weighted points for the action units can be visualized, see Fig. 8 (mean with 17 weights). This visualization can be found for all AUs in the supplementary material as well (mean with all weights). In most cases, the points with the highest weights correspond to the AU. In some cases however, almost all weights are high, which indicates that the used landmark detection approach is too noisy and therefore not suitable to detect the intensity of all AUs in all sequences.

## 4. Discussion

In this paper, we propose a method to generate one-dimensional time-varying feature responses from facial motion sequences based on feature points which describe the intensity of an expression. In four experiments we show the different applications of our approach. We evaluate the generated facial expression intensities of sequences from the BU4DFE dataset [19] with results from a state-of-the-art method [20] and show that we outperform them. Also, we evaluate our method by temporally aligning the sequences of BU4DFE. Compared to the commonly used method, our method generates superior feature responses to temporally align facial motion sequences. We use the ranking of weights generated with our method to find subclusters within each of the six prototypic emotions, revealing that the motion of the facial feature points used to perform one emotion differ between persons. We thereby show why clustering and classification of emotions with the six emotion labels is difficult. Finally, we create feature responses for specific AUs in the MMI database [11] and prove our feature can find the intensity of AUs impacting small areas, as well as multiple different AUs in a sequence.

In future work it would be interesting to use our feature for analysis of non-facial data, such as 3D full body motion capture [16, 17].